\documentclass[conference]{IEEEtran}

\usepackage[pdftex]{graphicx}
\usepackage{subfig}
\usepackage{rotating}
\usepackage{amsmath}
\usepackage{comment}
\usepackage{array}
\usepackage{caption}
\usepackage{setspace}
\usepackage{textcomp}
\usepackage{todonotes}
\usepackage{dingbat}
\usepackage{multirow}
\usepackage{url}
\newcolumntype{P}[1]{>{\centering\arraybackslash}p{#1}}

\newcolumntype{L}[1]{>{\raggedright\arraybackslash}p{#1}}
\newcolumntype{C}[1]{>{\centering\arraybackslash}p{#1}}
\newcolumntype{R}[1]{>{\raggedleft\arraybackslash}p{#1}}

\begin{document}

%\title{[WT] Procedural Content Generation via Quality Diversity}
\title{Procedural Content Generation\\ through Quality Diversity}

\author{
\IEEEauthorblockN{Daniele Gravina$^1$, Ahmed Khalifa$^2$, Antonios Liapis$^1$, Julian Togelius$^2$, Georgios N. Yannakakis$^1$}
\IEEEauthorblockA{
\textit{1: Institute of Digital Games, University of Malta}, Msida, Malta\\
\textit{2: Game Innovation Lab, New York University}, NY, USA\\
%daniele.gravina@um.edu.mt, ahmed.khalifa@nyu.edu, antonios.liapis@um.edu.mt,\\ julian@togelius.com, georgios.yannakakis@um.edu.mt
}
}
\IEEEoverridecommandlockouts
\IEEEpubid{\begin{minipage}{\textwidth}\ \\[12pt]
978-1-7281-1884-0/19/\$31.00 \copyright 2019 IEEE
\end{minipage}}
\maketitle

\begin{abstract}
Quality-diversity (QD) algorithms search for a set of good solutions which cover a space as defined by behavior metrics. This simultaneous focus on quality and diversity with explicit metrics sets QD algorithms apart from standard single- and multi-objective evolutionary algorithms, as well as from diversity preservation approaches such as niching. These properties open up new avenues for artificial intelligence in games, in particular for procedural content generation. Creating multiple systematically varying solutions allows new approaches to creative human-AI interaction as well as adaptivity. In the last few years, a handful of applications of QD to procedural content generation and game playing have been proposed; we discuss these and propose challenges for future work.
\end{abstract}


\IEEEpeerreviewmaketitle

\section{Introduction}\label{sec:introduction}
Since \emph{ROGUE} (Toy and Wichman, 1980) and \emph{Elite} (Acornsoft, 1984) in the 1980s, certain genres of digital games have relied on algorithmic processes to generate content such as levels, weapons, personalities, quests, etc. Throughout its long history, procedural content generation (PCG) has aimed to provide content that is playable, of a high quality, and yet different from other content that came before or after.  On the one hand, most game content need to satisfy certain minimal criteria on playability (such as the exit in a dungeon being reachable by the player) while they also need to be entertaining and challenging (which are softer and often subjective quality dimensions). Content which do not satisfy these criteria of quality can break the gameplay explicitly or implicitly, resulting in a poor player experience. On the other hand, games that rely on PCG to produce fresh content promise that every playthrough, opened chest, or visited settlement will be different. Players expect novel and unseen content at every possible moment, and can swiftly turn against a game where the variation in generated content is low or cosmetic. The backlash against \emph{No Man's Sky} (Hello Games, 2016) was in no small part due to the lack of perceptible variety in the generated worlds \cite{maiberg2016nomanssky}. While low-quality generated content can at best affect players' enjoyment and at worst make a game unbeatable, generated content with insufficient diversity can lead to fatigue and dejection.
%\todo[inline]{Daniele: computatioal creativity addition}
If we look at the large body of research in computational creativity, we can see that value and novelty are vital criteria to evaluate an artifact as creative \cite{ritchie2007some}. While value is commonly targeted in traditional PCG, novelty---or more broadly, game content diversity---is not. %; however, it has been recognized as a desirable property for PCG algorithms \cite{preuss2014gooddiverse}. 

In sum, it has been established that many PCG problems require both \emph{quality} and \emph{diversity} of the generated content \cite{preuss2014gooddiverse}. This poses a challenge for many existing PCG methods, which are often forced to make a trade-off between these two requirements. In this paper, we argue that recent advancements in search methods allow us to overcome this problem and create PCG algorithms that emphasize quality and diversity simultaneously. This has the potential to significantly increase the scope of what PCG can do in games.

Quality-Diversity (QD) algorithms are a novel family of evolution-like algorithms that simultaneously maintain the quality and diversity of their solutions by rewarding  divergence (as novelty or surprise of the artifacts being generated) while maintaining control of the solutions' quality through hard constraints or local competition among individuals with similar behavioral traits. We propose, therefore, procedural content generation through quality-diversity (PCG-QD) as a subset of search-based procedural content generation \cite{togelius2011search} which is perfectly suited for generating content autonomously (as it can produce a large set of diverse and high-quality artifacts in one run, even in search spaces which are not well-defined) or with a human designer (as it can explain and express its artifacts' desirable properties). This paper presents the components of quality-diversity algorithms, identifies the strengths of PCG-QD over popular alternatives, surveys recent work in this vein and attempts to map out the road ahead.

\section{Quality Diversity Approaches}\label{sec:algorithms}

%This section reviews the idea behind the concept of quality diversity and main quality diversity approaches proposed literature. 
Inspired by the extreme diversity of high-performing creatures found in nature, the \emph{quality diversity} (QD) paradigm \cite{pugh2016quality} has the goal of finding the largest possible set of diverse and high-quality solutions in one run.
%\todo[inline]{Daniele: multimodal optimization addition}
The search for multiple solutions to a problem is a long-standing challenge in evolutionary computation. In multimodal optimization, for instance, the aim is to find all the local optima of a function by employing niching methods and genotypic diversity mechanisms \cite{preuss2015multimodal}. However, only searching for local optima does not guarantee the diversity of the solutions, as (behaviorally) diverse solutions may show the same performance measure \cite{cully2017quality}. 
Quality diversity, instead, stresses the importance of searching for diverse solutions \emph{first} and then maximize their quality.
Indeed, QD draws inspiration from the idea of rewarding divergence to find the necessary steps towards high performing areas of the search space. In divergent search, artificial evolution is not guided by a fitness tied to the ultimate objective of the problem, but instead rewards directly the diversity of solutions, based on notions such as novelty~\cite{lehman2011abandoning}, surprise~\cite{gravina2016surprise} or curiosity~\cite{stanton2016curiosity}.
QD combines the divergent properties of divergent search with localized convergence, as will be discussed below.

\subsection{Divergence Components}
As discussed above, QD is based on purely divergent search approaches and thus its core drive is to maintain and reward diversity in a population. There are several ways in which this can be achieved in QD algorithms available today: 
%This section talks about the different ways that current known QD algorithms use to maintain diversity between individuals.

\subsubsection{Behavior Space Distance}
Diversity in all known divergence search approaches is measured based on the distance of two individuals in a \emph{behavioral space}. The premise is to have a distance function that can give us a single value that determines how different the behavior of the individual is from the rest of individuals. Unlike other diversity preservation mechanism such as niching \cite{preuss2012niching}, divergence is judged on the behavioral space rather than on genotypic similarity. The rest of individuals can be the current population and an archive of past individuals, as in Novelty Search~\cite{lehman2011abandoning}, a predicted snapshot of the population as in Surprise Search~\cite{gravina2016surprise}, or explored areas in this individual's lifetime as in Curiosity Search~\cite{stanton2016curiosity}.

\subsubsection{Behavior Space Partitioning}
Diversity can also be enforced by partitioning the behavior space using $N$ different dimensions of behavior, where each dimension is discretized and stored as a grid or a \emph{map} of cells. Each cell corresponds to a different area in the behavior space with different properties; this cell contains all individuals that have a certain behaviors. Partitioning could be uniform, as in MAP-Elites~\cite{mouret2015illuminating}, or based on the distribution of the population, as in MAP-Elites with Sliding Boundaries~\cite{fontaine2019mapping}. A big difference between the use of a distance function and partitioning the space is that partitioning allows designers to control the granularity of the space.

\subsection{Quality Components}
Current QD algorithms have a number of different strategies for pushing evolutionary search towards improving the quality of candidate solutions:

\subsubsection{Local Competition}
A simple way to enhance the quality of the solution is using local competition between individuals within the same niche. 
%Enforcing a local competition between individuals within the same niche can enhance the quality of the solutions. 
To apply this method, a distance metric must be specified to discern which candidates an individual compares itself with. This distance can be the same as the behavior space distance, or could be based on the behavior space partitioning where competing individuals would share the same cell. 

\subsubsection{Constraints}
Another way to ensure quality is through a set of constraints that each individual tries to satisfy. These constraints are usually ``hard constraints'' (e.g. a level can be completed or not), dividing the population into \emph{infeasible} individuals and \emph{feasible} individuals. In most of the  previous work~\cite{liapis2015constrained,liapis2013sentient,liapis2013transforming,gravina2016constrained}, infeasible individuals are not considered in terms of their diversity. However, in some cases the diversity is always maintained regardless of constraint satisfaction~\cite{khalifa2018talakat,khalifa2019intentional}.

\subsection{Algorithms}
This subsection discusses the different QD algorithms used in previous work, even if they have not been applied to games.

\subsubsection{MAP-Elites (ME)}
MAP-Elites ``illuminates'' the search space~\cite{mouret2015illuminating}, highlighting which attributes of the solutions can contribute to their performance. MAP-Elites combines behavior space partitioning to maintain diversity with local competition to improve the quality by only maintaining the best individual in each cell.
The algorithm starts by generating a random set of solutions, which are evaluated in terms of quality and behavioral properties; the latter are used to place this individual in a cell in the partitioned behavior space (feature map). % The appropriate cell for each solution is calculated based on the recorded behavior. 
If the chosen cell is empty, the solution is stored in the map; if the cell is occupied, a local competition occurs and the best individual between the two is kept. After this initialization procedure, the stored solutions are uniformly selected to generate a new individual via crossover and/or mutation.
The new solution is evaluated and may be stored in the chosen cell if its performance is better than the solution (elite) currently stored there.
These phases---selection, mutation, evaluation and replacement---are repeated for a number of evaluations or until the grid reaches a desired coverage.

\subsubsection{MAP-Elites with Novelty Search (ME-NOV)}
To bias exploration towards underrepresented solutions in MAP-Elites, \cite{pugh2016quality} used novelty search to select individuals in the feature map. The resulting QD algorithm selects elites to generate offspring proportionally to their novelty score~\cite{lehman2011abandoning}. Offspring are added to the novelty archive, which is used to compute the novelty for each elite stored in the grid. Choosing an appropriate behavioral distance for novelty can bias the areas explored by the algorithm, and a number of variants have been proposed such as combining two feature maps for two behavioral characterizations~\cite{pugh2016quality} or searching for novelty in a dimension orthogonal to the features used for space partitioning~\cite{gravina2019blending}.

\subsubsection{MAP-Elites with Sliding Boundaries (MESB)}
Rather than use a predetermined cell size, MESB \cite{fontaine2019mapping} recomputes the boundaries of the MAP-Elites feature map based on the current distribution of individuals. For every $\lambda$ individuals generated, the boundaries are recomputed uniformly based on the percentage marks of the distribution along each feature considered. MESB can thus dynamically adapt to unequal distributions, better representing the underlying distribution of the high-performing individuals in the feature space.

\subsubsection{Constrained Novelty Search (CNS)}
This QD approach combines the Feasible-Infeasible Two-Population Genetic Algorithm (FI-2Pop)~\cite{kimbrough2008feasible} with novelty search, in order to maximize the behavioral space distance of individuals which satisfy certain constraints on quality. CNS maintains two populations, one with only infeasible individuals and one with only feasible. The feasible population evolves to maximize novelty, via novelty search~\cite{lehman2011abandoning}, while the infeasible population evolves towards minimizing the distance from feasibility (FINS) or again to maximize novelty (FI2NS)~\cite{liapis2015constrained}.
The generated offspring of two populations can migrate from one population to the other (based on feasibility), which increases the chance %of a serendipitous exchange of genetic materials between the two populations and the possible
to discover stepping stones toward multiple high-quality solutions. 

\subsubsection{Constrained Surprise Search (CSS)}
Similar to constrained novelty search, this QD algorithm uses unexpectedness as the diversity measure for evaluating feasible individuals in a FI-2pop GA. As in the above variant, CSS is built on the FI-2Pop genetic algorithm, where the infeasible population  minimizes the distance from feasibility and the feasible population maximizes surprise through surprise search~\cite{gravina2016surprise,gravina2019qualitydiversity}.

\subsubsection{Constrained MAP-Elites (CME)}
Combining the constraint-satisfaction capabilities of FI-2Pop with MAP-Elites, constrained MAP-Elites (CME)~\cite{khalifa2018talakat,khalifa2019intentional} maintains two populations for each cell of the archive, one containing infeasible individuals and one containing feasible individuals. As in the original formulation of FI-2Pop~\cite{kimbrough2008feasible}, the feasible population maximizes the predefined fitness, while the infeasible population minimizes the distance from feasibility. Every chromosome is stored in the corresponding cell and population, and the algorithm benefits from the unique features of illumination and constraint satisfaction.

\subsubsection{Novelty Search with Local Competition (NS-LC)}
Combining optimization towards a behavior space distance with quality control via local competition, NS-LC~\cite{lehman2011creatures} applies a multi-objective approach to maximize both the novelty score and a local competition objective which pushes the individual to outperform others in its niche. Novelty search~\cite{lehman2011abandoning} rewards solely the solutions' novelty and ignores the objective of the problem, selecting solutions based on their behavior space distance from other solutions in the population as well as past novel solutions stored in the \emph{novelty archive}. Local competition rewards those individuals that perform better compared to their $K$ nearest neighbours. Local competition ensures a localized convergence considering only the nearest neighbors in the current generation and the novelty archive, creating a local selection pressure across the entire search space.

\subsubsection{Surprise Search with Local Competition (SS-LC)}
Similarly to NS-LC, this QD algorithm uses unexpectedness rather than novelty to select individuals in one objective, and local competition as another objective in a multi-objective fashion. Unexpectedness in surprise search \cite{gravina2016surprise} is measured as deviation from predictions made on past generations, which offers a different measure of divergence than novelty which rewards unseen solutions~\cite{gravina2019qualitydiversity}. However, surprise search can also be combined with novelty search and local competition, which leads to good performance in highly deceptive domains~\cite{gravina2019qualitydiversity}.

\section{Why Quality Diversity?} \label{sec:why_qd}

This section highlights what makes PCG-QD an important contribution to the panorama of PCG research. Since all QD approaches surveyed in Section~\ref{sec:algorithms} are based on artificial evolution, PCG-QD is a subset of search-based PCG~\cite{togelius2011search}. However, several key properties of QD algorithms make them better suited than `typical' SBPCG approaches. Moreover, PCG-QD is compared with popular approaches for generating games in academia and in the industry, namely machine learning (PCGML)~\cite{summerville2018procedural} and constructive algorithms~\cite{shaker2016constructive} respectively.

\subsection{Generative Efficiency}\label{sec:why_efficiency}
The ability of QD approaches to produce a large set of high-quality solutions which exhibit diverse behaviors in one run makes it surprisingly efficient when a broad variety of content is needed. By comparison, constructive algorithms may be computationally fast but output a single artifact; moreover, re-running the algorithm does not ensure that the new output will be particularly different than previous ones. This lack of originality in generated content has been lamented in several games such as \emph{No Man's Sky} (Hello Games, 2015). PCGML similarly outputs a single artifact, and the lack of diversity from one run to the next is even more obvious as the generated content attempt to explicitly follow the same patterns. Traditional SBPCG approaches, while evolving a population of artifacts, are usually interested in the fittest individual at the end of an evolutionary run. Other generative approaches which are fast in producing a feasible individual, such as declarative programming~\cite{smith2011designspace}, have no way of controlling the diversity of their output. Therefore, while QD can be computationally heavy, a single run can output a vast corpus of high-quality content; this removes the burden of re-running and post-generation assessment of both quality and diversity.

\subsection{Fitness-Free Search}\label{sec:why_unbiased}

The underlying assumption of search-based approaches~\cite{togelius2011search} is that by defining the fitness function, we can generate content of high quality based on the designer's definition of a \emph{quality measure}. However, defining an effective objective function is not an easy task, exacerbated by known problems such as deceptive fitness landscapes~\cite{Goldberg1987a, lehman2011abandoning}. 
Furthermore, when the objective of the problem to solve depends on subjective criteria, as is often the case for games, it is difficult to formalize the value to optimize~\cite{csikszentmihalyi1992optimal}. 
It is widely recognized that designing a fitness function that incorporates (a) subjective and (b) multiple criteria is a challenging, especially considering that games are multifaceted~\cite{liapis2014gamecreativity}. 
Due to their multifaceted nature, an algorithm designer might need to consider both non-functional properties, such as the aesthetic properties and functional properties, e.g., playable levels.
While in the literature several solutions have addressed this~\cite{togelius2011search,preuss2014gooddiverse}, we argue that quality-diversity fits particularly well as an answer to this problem. Quality-diversity can consider \emph{more than one} dimensions of interest and at the same time explores the fitness landscape based on local rather than global competition in terms of a fitness function. Combined with the explainability of QD approaches (see Section \ref{sec:why_explainability}), this can highlight potential biases in the chosen fitness function.

\subsection{Online Expressivity Analysis}\label{sec:why_expressivity}

A unique feature of quality-diversity as a procedural content generator is the \emph{online expressivity analysis} granted as a byproduct of the search for highly diverse and high-quality solutions.
Expressivity analysis~\cite{smith2010analyzing} is defined as the analysis of the output in terms of styles and variety of artifacts generated by the chosen approach, which can highlight biases of the generator towards specific types of content. While all generators can produce (with multiple re-runs) the large set of artifacts required to perform an expressivity analysis, as noted in Section~\ref{sec:why_efficiency} the QD approaches produce such sets in a single run. More importantly in this context, the diversity components of these QD algorithms perform an expressivity analysis \emph{during} evolution (i.e. online). Moreover, QD algorithms which explicitly optimize for behavioral distance explicitly attempt to increase the expressivity of the generator by searching under-explored niches.

\begin{figure}
    \centering
    \includegraphics[width=0.7\linewidth]{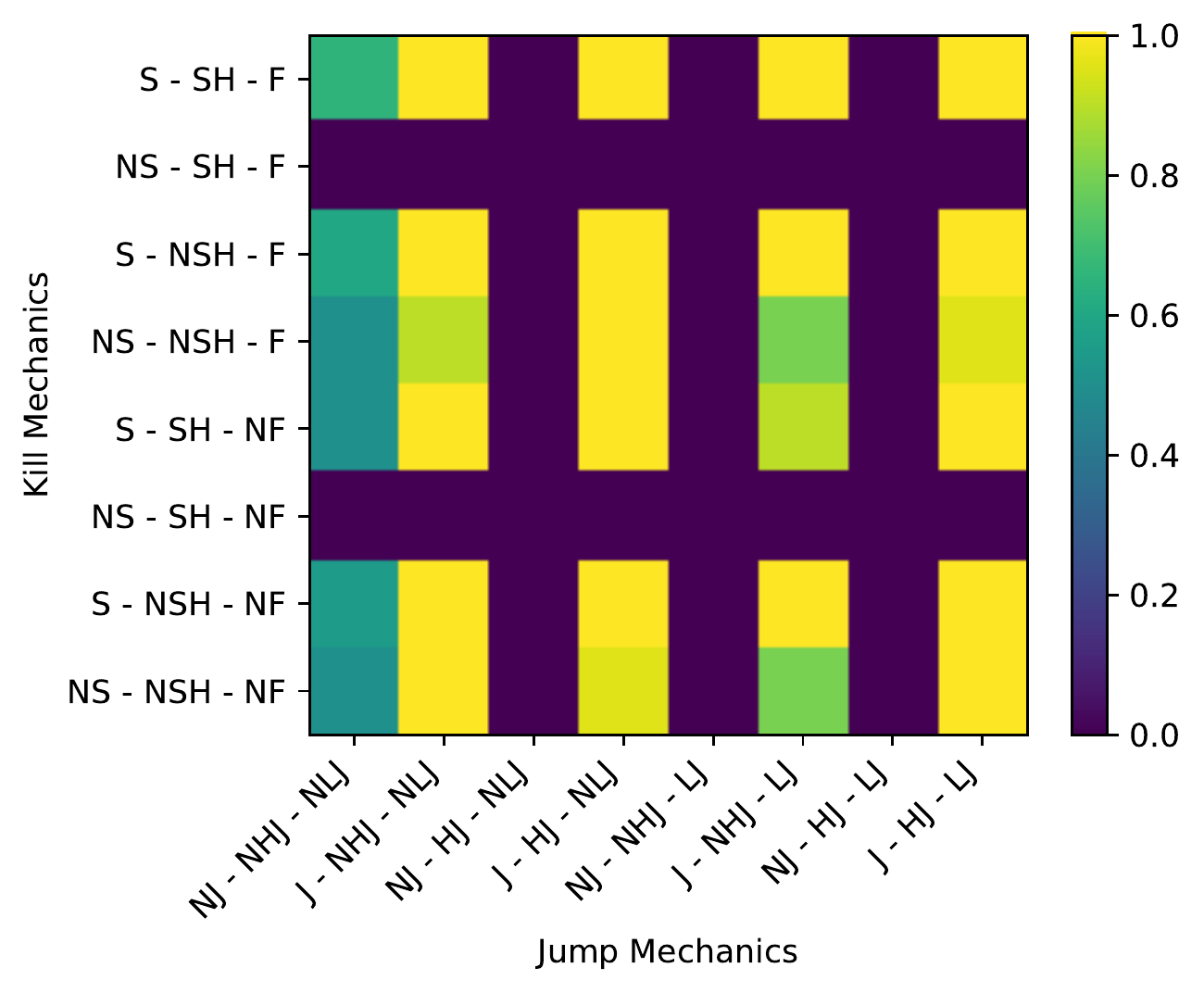}
    \caption{Expressivity as a heatmap of six binary features (combinations of used mechanics) for Mario scenes in \cite{khalifa2019intentional}.} % where each feature is labeled with two symbols: \textit{X} and \textit{NX}. \textit{X} if a specific mechanic happened in the playthrough and \textit{NX} otherwise. X-axis has 3 different jumps mechanics: \textit{J} if the player had jumped, \textit{LJ} is long jump, and \textit{HJ} is high jump. Y-axis has 3 different kill mechanics: \textit{S} is stomp kill, \textit{SH} is shell kill, and \textit{F} is fall kill.}
    \label{fig:why_expressivity}
\end{figure}

Figure~\ref{fig:why_expressivity} shows an example of the covered space from the work of Khalifa \emph{et al.}~\cite{khalifa2019intentional} on generating scenes in \emph{Super Mario Bros} (Nintendo, 1985), elaborated in Section~\ref{sec:cases}. The figure shows that no levels are generated for some areas of the search space because the divergence characterizations chosen are dependent on each other. For example: a player can not kill an enemy in \emph{Super Mario Bros} without at least jumping. This problem of behavior characterization (BC) dependency will be discussed in more detail in Section~\ref{sec:outlook}.

\subsection{Human-Machine Co-Creation}\label{sec:why_mixedinitiative}

Game development often involves design iterations that can completely change the objectives and design priorities. When game developers work alongside an AI-assisted tool~\cite{smith2011tanagra,liapis2013sentient}, the tool's ability to illuminate the space with multiple different and good solutions can help designers identify new designs or to perfect generated content based on their current priorities. QD approaches are able to efficiently (see Section \ref{sec:why_efficiency}) produce a diverse set of high-quality content, and give a designer control to adjust both the criteria of quality and the behavioral characterization of the artifacts. This makes QD approaches especially effective tools for mixed-initiative content design, and can foster their users' creativity with expected or unexpected but always high-quality suggestions~\cite{liapis2016canmixedinitiative}.

\subsection{Explainability}\label{sec:why_explainability}

Explainable AI for Designers~\cite{zhu2018explainable} is an important research area that aims to aid the game designer in understanding AI algorithms applied to games. Such explainability is useful during co-creative tasks (see Section~\ref{sec:why_mixedinitiative}) but also for debugging purposes or when the generated artifact is used in another design iteration. Since QD approaches such as MAP-Elites \emph{illuminate} the search space and visualize it as a feature map, this can help developers explore and understand the generator's output \cite{cook2016danesh}. Explainability in this vein is tied to the online and embedded expressivity analysis of these algorithms (see Section \ref{sec:why_expressivity}). As with any evolutionary algorithm in the SBPCG family, QD approaches can explain the origins of the artifact by showing the lineage of any individual (e.g.,~in \cite{hastings2009garjournal,secretan2011picbreeder}). Strengthening this latter form of explanation, the fact that PCG-QD is efficient in producing a set of good and diverse artifacts in one run (see Section \ref{sec:why_efficiency}) strengthens this lineage visualization as the common ancestors of different sets of high-performing content can be shown. These different visualizations can help the designer group individuals and select those with the desired features among the many individuals generated by the algorithm.

\section{Cases of Quality Diversity in PCG}\label{sec:cases}

\begin{table*}
\centering
\small
\begin{tabular}{|c|c c | c c| P{0.21\linewidth} | P{0.21\linewidth} | c |}
\hline
\multirow{3}{*}{\textbf{Algorithm}} & 
\multicolumn{4}{c|}{\textbf{Components}} & 
\multicolumn{2}{c|}{\textbf{Characterization}} &
\multirow{3}{*}{\textbf{Artifact}} \\
\cline{2-7}
& \multicolumn{2}{c|}{Divergence}
& \multicolumn{2}{c|}{Quality} 
& \multirow{2}{*}{Divergence}
& \multirow{2}{*}{Quality}
&
\\
\cline{2-5}
&  D
&  P
&  LC 
&  C 
& 
&
&
\\

\hline
MAP-Elites & - & \checkmark &  \checkmark & - & DNN Output & DNN Confidence & 2D and 3D objects~\cite{nguyen2015innovation, lehman2016creative} \\
\hline
MESB       & - & \checkmark &  \checkmark  & - & Mana Distribution & Health Difference & Hearthstone Decks~\cite{fontaine2019mapping}\\
\hline
\multirow{3}{*}{CME} & - & \checkmark & \checkmark & \checkmark & Playthrough Properties & Validity and Playability & Bullet-Hell Scripts~\cite{khalifa2018talakat} \\
& - & \checkmark & \checkmark & \checkmark  & Triggered Mechanics & Playability and Simplicity & Mario Scenes~\cite{khalifa2019intentional} \\
& - & \checkmark & \checkmark & \checkmark  & Linearity, Simmetry, Similarity, and Patterns & Playability, Room properties, and Design patterns & Dungeons~\cite{alvarez2019empowering} \\
\hline
\multirow{3}{*}{CNS}  & \checkmark & - & - & \checkmark  & Visual Diversity & Playability & Map Sketches~\cite{liapis2015constrained, liapis2013sentient} \\
& \checkmark & - & - & \checkmark & Visual Diversity & Believability & Arcade-Style Spaceships~\cite{liapis2016arcade}\\
& \checkmark & - & - & \checkmark  & DNN Latent Space & Believability & 2D Spaceship Hulls~\cite{liapis2013transforming}\\
\hline
CSS & \checkmark & - & - & \checkmark & Map Locations & Balance and Playability  & FPS Weapons~\cite{gravina2016constrained}\\
\hline
NS-LC & \checkmark & - & \checkmark & - & Block Presence & Complexity & Minecraft-like Structures~\cite{soros2017voxelbuild} \\ 
\hline
\end{tabular}
\caption{Cases of PCG through Quality Diversity. In the components column, \emph{D} stands for Behavior Space Distance, \emph{P} for Behavior Space Partitioning,  \emph{LC} for Local Competition and \emph{C} for Constraints.}
\label{tab:cases}
\end{table*}

%This taxonomy tries to highlight the differences and commonalities between the approaches proposed in the previous works and the missing combinations of these different basic components.
%In particular, we take in consideration three well-known approaches used in Quality-Diversity: MAP-Elites \cite{mouret2015illuminating}\, Multi-objective and FI-2Pop \cite{kimbrough2008feasible}.

This section discusses previous work in PCG-QD, highlighting the differences and commonalities between the different algorithms used.
Table~\ref{tab:cases} examines each case in terms of five components: the QD algorithm, components of Section~\ref{sec:algorithms}, the quality and diversity characterizations, and the type of artifact produced.

\subsection{Generation of 2D and 3D Objects}
 
MAP-Elites was used to generate 2D images~\cite{nguyen2015innovation} and 3D objects~\cite{lehman2016creative}. Similar to DeLeNoX~\cite{liapis2013transforming}, this work combines quality-diversity
%\todo[inline]{julian: Shouldn't we mention that this is essentially a re-invention of DeLeNoX, as Jeff himself admitted? AL: I mean, do we gain something by doing this?}
%julian Shouldn't we mention that this is essentially a re-invention of DeLeNoX, as Jeff himself admitted?
search and machine learning. A Deep Neural Network (DNN) is first trained on classifying real-world images and then combined with MAP-Elites to generate 2D images or 3D objects. The classification output of the DNN distinguishes between 1000 different classes of images, which is used as diversity characterization of the input.
%where the employed fitness is the confidence of the DNN. 
Specifically, MAP-Elites uses the classification output of the DNN as a partition of the search space and tries to optimize every bin based on the confidence of the DNN.

\subsection{Generation of Bullet Hell Scripts}

\begin{figure}[!tb]
	\centering
	\subfloat{
		\includegraphics[width=0.3\linewidth]{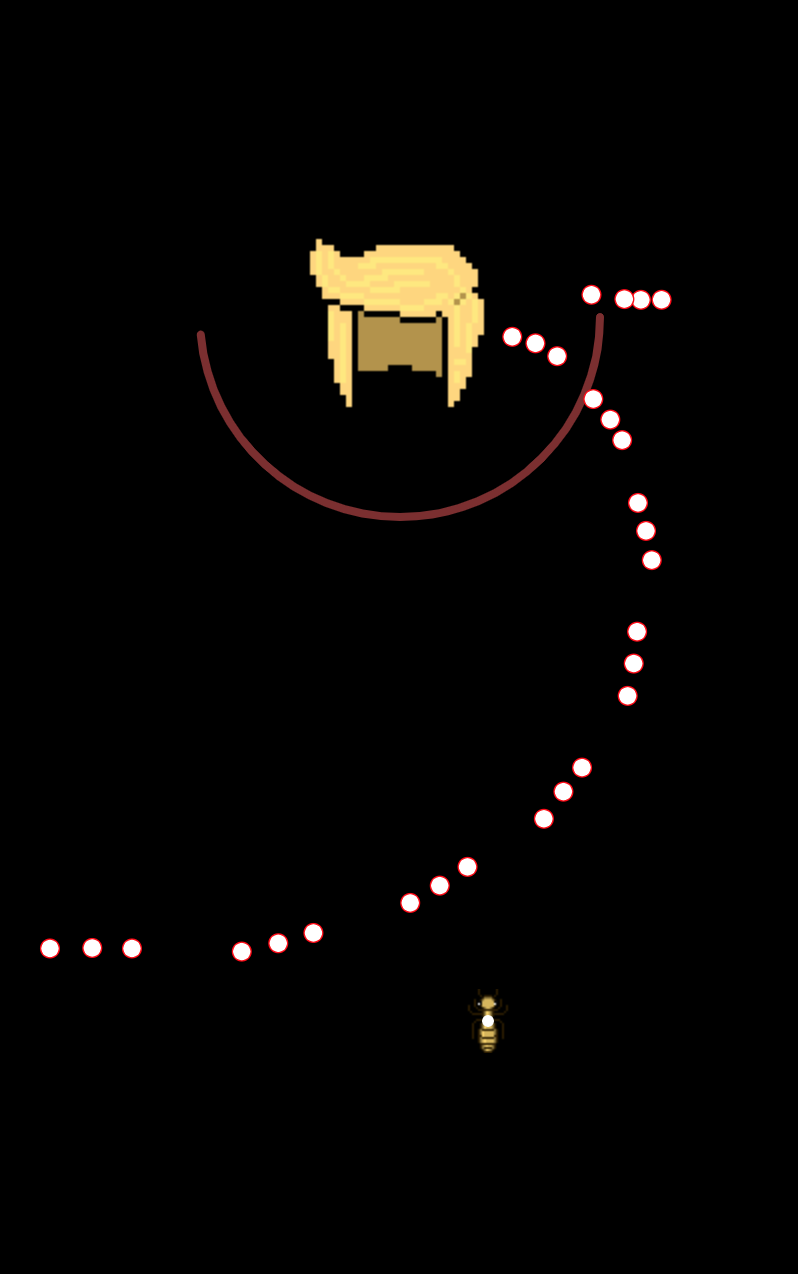}
	}\hfill
	\subfloat{
		\includegraphics[width=0.3\linewidth]{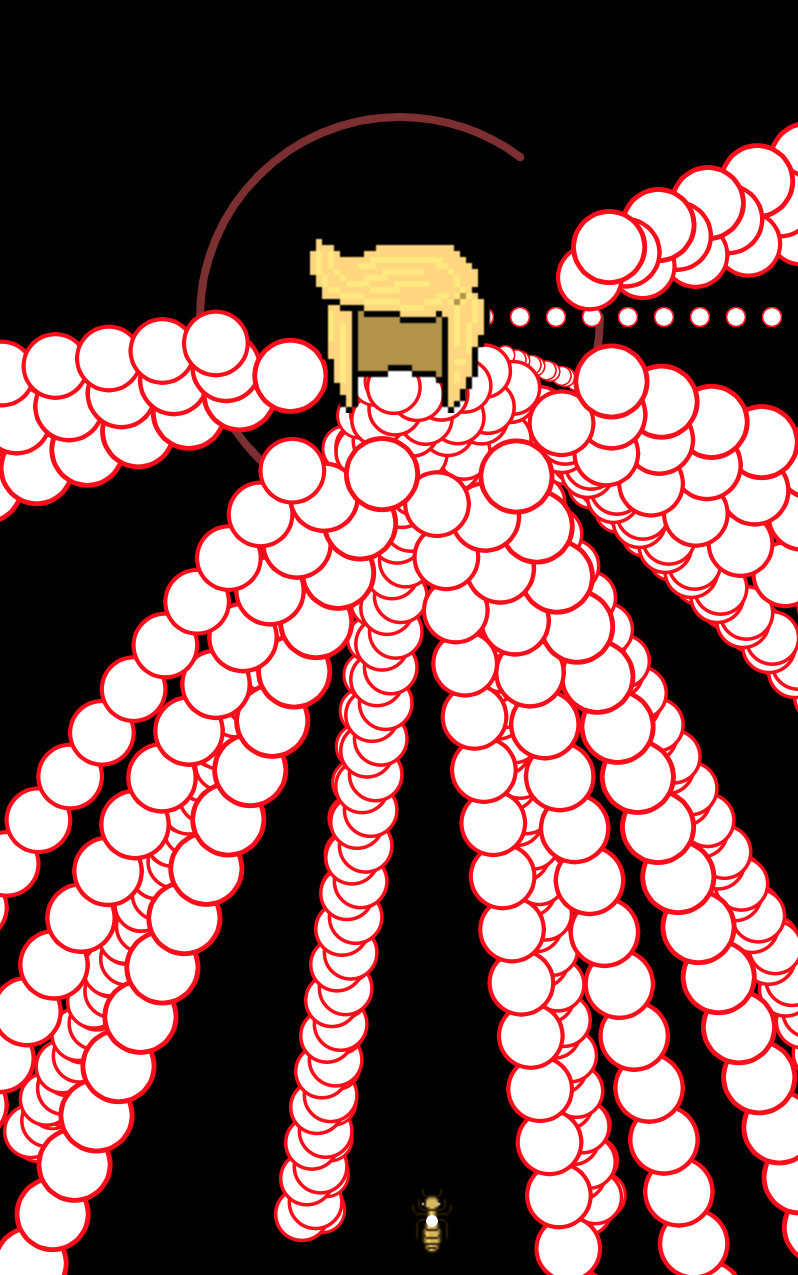}
	}\hfill
	\subfloat{
		\includegraphics[width=0.3\linewidth]{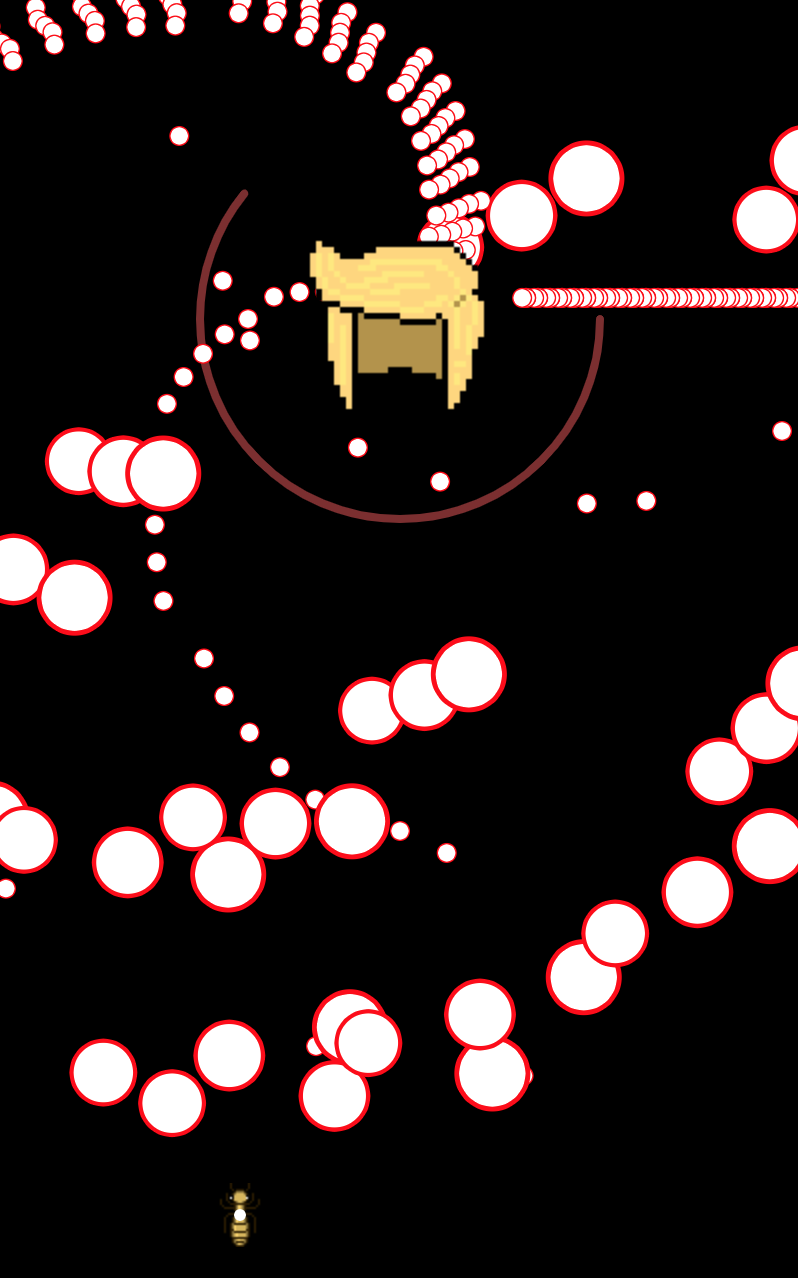}
	}
	\caption{Bullet-hell level created with constrained MAP-Elites. Figures reproduced with permission from the authors~\cite{khalifa2018talakat}.}
	\label{fig:talakat_levels}
\end{figure}

A constrained version of MAP-Elites was used to generate levels for bullet hell games~\cite{khalifa2018talakat}. The generated levels are represented as a sequence of events in the Talakat description language~\cite{khalifa2018talakat}. The algorithm tries to evolve valid scripts while making sure that the level is playable using an A* algorithm. The authors use features of the agent and the level (agent entropy, agent risk, and bullet distribution) as dimensions of the feature map. Figure~\ref{fig:talakat_levels} shows three different generated levels from this experiment from different areas in the map.

\subsection{Generation of Mario Scenes}

\begin{figure}[!tb]
	\centering
	\subfloat{
		\includegraphics[width=0.3\linewidth]{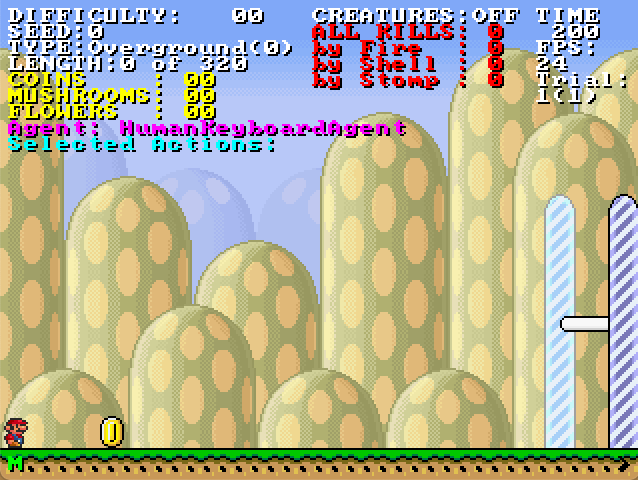}
	}\hfill
	\subfloat{
		\includegraphics[width=0.3\linewidth]{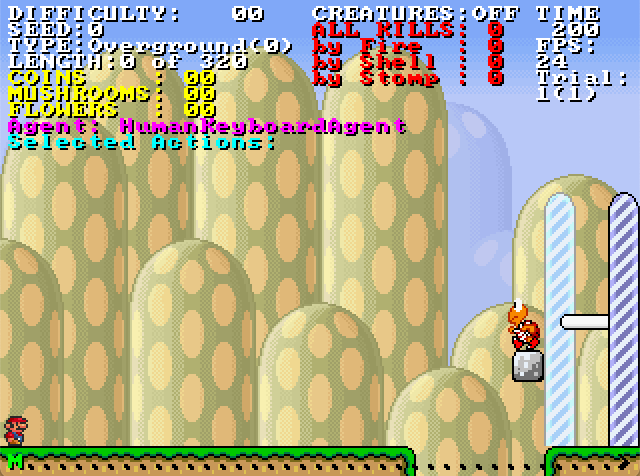}
	}\hfill
	\subfloat{
		\includegraphics[width=0.3\linewidth]{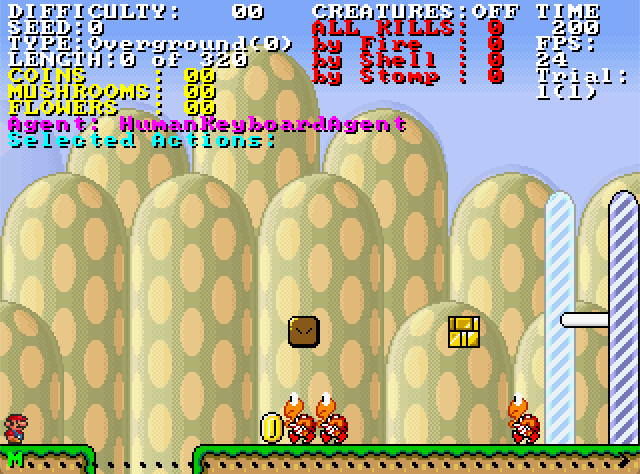}
	}
	\caption{Super Mario Bros scenes created with constrained MAP-Elites. Figures reproduced with permission from the authors~\cite{khalifa2019intentional}.}
	\label{fig:mario_levels}
\end{figure}

Constrained MAP-Elites was used to generate small sections of levels (scenes) for \emph{Super Mario Bros}~\cite{khalifa2019intentional}. The algorithm tries to find the simplest playable scenes, using an A* agent to check for playability. The algorithm uses the triggered game mechanics during the playthrough as binary dimensions for the map (the mechanic is either fired during the playthrough or not) to make sure the generated scenes target different game mechanics. Eight different mechanics were used, thus creating a feature map of 256 cells. The algorithm was able to generate levels in 100 cells on average (see Fig.~\ref{fig:why_expressivity}), ranging from levels where no mechanics were fired to levels where all mechanic were fired. 
Figure~\ref{fig:mario_levels} shows three different scenes with different degrees of fired mechanics ranging from no fired mechanics on the left to five mechanics being fired on the right.

\subsection{Generation of Hearthstone Decks}

MAP-Elites with Sliding Boundaries~\cite{fontaine2019mapping} was used to generate decks for the online card game \emph{Hearthstone} (Blizzard, 2014).
The main goal is to generate high-performing decks that vary in their \emph{mana} distribution curves (mana is the cost required to play a card). 
In order to do so, the fitness is computed as the difference between the two players' health in 200 games; diversity is based on the distribution of the mana curve, codified as the average and variance of the mana distribution in the population.
%Figure~\ref{fig:hearthstone} shows the mana distributions generated for different cases.

\subsection{Generation of Map Sketches}

\begin{figure}[!tb]
	\captionsetup[subfigure]{labelformat=empty}
	\centering
	\includegraphics[width=0.8\linewidth]{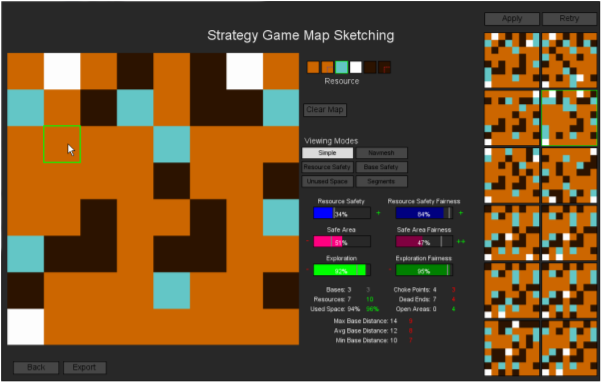}
	\caption{A screenshot taken from the Sentient Sketchbook tool. Figure reproduced with permission from the authors~\cite{liapis2013sentient}.}
	\label{fig:sentient_sketchbook}
\end{figure}

%Mixed-initiative PCG-QD has been explored in \cite{liapis2013sentient, liapis2015constrained}.
Sentient Sketchbook~\cite{liapis2013sentient} is a mixed-initiative game design tool where the user can design the level of the game and receive in real-time feedback on playability constraints, balance, etc. PCG-QD has the role of inspiring the user with level suggestions through genetic search. In particular, constrained novelty search~\cite{liapis2015constrained} is used to produce suggestions using the designer's sketch as an initial seed: a combination of FI-2Pop and novelty search guarantees feasible and highly diverse suggestions. Feasibility tests whether the level is playable (e.g. all resources tiles are reachable by every player's base), while diversity is computed as the number of tiles which are different. Figure~\ref{fig:sentient_sketchbook} shows a screenshot of a design session.

\subsection{Generation of Weapons}

\begin{figure}[!tb]
	\captionsetup[subfigure]{labelformat=empty}
	\centering
	\subfloat{
		\includegraphics[width=0.45\linewidth]{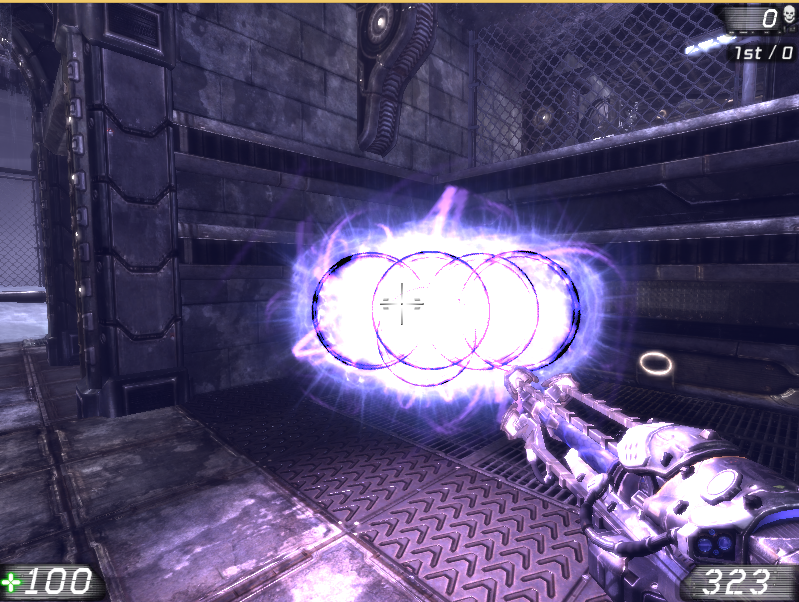}
	}
	\subfloat{
		\includegraphics[width=0.45\linewidth]{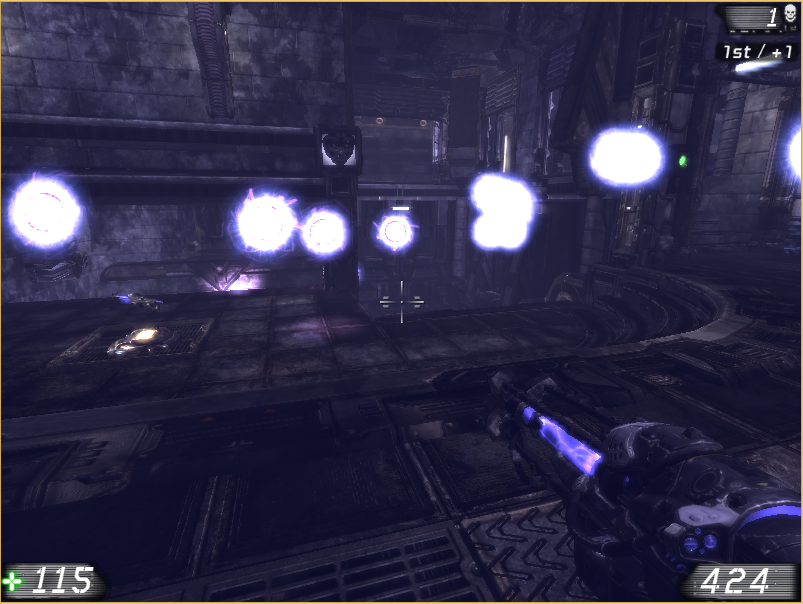}
	}
	\caption{Two example weapons created by constrained surprise search. Figures reproduced with permission from the authors~\cite{gravina2016constrained}.}
	\label{fig:weapons_surprise}
\end{figure}

A constrained approach to generate a number of diverse and functional weapons for competitive First Person Shooter games is introduced in~\cite{gravina2016constrained}. These weapons need to be usable and balanced but also exhibit surprising behaviors. In order to accomplish that, a constrained surprise search algorithm is devised to generate weapons that are feasible (balanced and usable) and surprising in their behaviors. In particular, two populations are employed, one where infeasible weapons are optimized towards feasibility and another population where the feasible weapons are rewarded based on their unexpectedness. Surprise is computed based on the death location of the agents used to simulate gameplay with the evolved weapons. A heatmap is computed in every generation based on the coordinates of these death locations and a prediction is made via linear regression of the heatmaps computed in the last two generations. Surprise is then computed based on the difference between the individual's death locations and the predicted heatmap.
As an example, Figure~\ref{fig:weapons_surprise} shows a pair of generated weapons through this QD approach.

\subsection{Generation of Spaceships}

% \begin{comment}
% \begin{figure}[!tb]
% 	\captionsetup[subfigure]{labelformat=empty}
% 	\centering
% 	\includegraphics[width=0.5\linewidth]{./images/spaceships}
% 	\caption{Sample spaceships generated with DeLeNoX. The images are produced from different iterations during the exploration phase. Figure reproduced with permission from the authors~\cite{liapis2013transforming}.}
% 	\label{fig:spaceships}
% \end{figure}
% \end{comment}

Constrained novelty search has been used extensively for generation of spaceships' visuals, driven by constraints on visual quality (such as all shapes being connected) and by divergence either based on pre-authored visual properties~\cite{liapis2016arcade} or based on emergent visual patterns. For the latter, DeLeNoX~\cite{liapis2013transforming} generated increasingly diverse spaceships based on a latent representation and the exploration capabilities of novelty search. Specifically, DeLeNoX alternates two key algorithmic phases, exploration and transformation. During the exploration phase, constrained novelty search generates many diverse but feasible two-dimensional spaceships; during the transformation phase, an autoencoder tries to compress all generated solutions in a low-dimensional latent space. The subsequent exploration phase uses the autoencoder's latent space as behavior characterization, and thus the diversity measure is continuously adapted in every cycle of DeLeNoX. 
%Fig. \ref{fig:spaceships} shows a sample of the generated spaceships.

\subsection{Generation of Minecraft-like Structures}

Novelty search with local competition~\cite{lehman2011creatures} was used to generate a number of diverse block structures for the popular video-game \emph{Minecraft} (Mojang, 2011) in ~\cite{soros2017voxelbuild}. The block structures are evolved via Artificial Neural Networks, which are in turn evolved via NS-LC. 
In particular, structures' quality is based on their complexity (net number of placed blocks times the height of the tallest block), while the behavior characterization is a binary vector specifying whether there is a block at each location in the simulation.

\subsection{Generation of Dungeons}
%\todo[inline]{Daniele: Addition}
An interactive variant of constrained MAP-Elites (CME) is proposed in \cite{alvarez2019empowering} to blend quality-diversity with the mixed-initiative system Evolutionary Dungeon Designer (EDD). EDD is a design tool capable of generating dungeons for adventure games. In this work, CME is augmented with mixed-initiative capabilities and tested in EDD with different pairs of dimension of interest: symmetry and similarity, number of meso-patterns, number of spatial patterns, and linearity. As in the original CME implementation, the infeasible population minimizes the feasibility constraint (playability) while the feasible population maximizes a weighted sum of inventorial aspects of the room and the spatial distribution of design patterns. The user can choose two dimensions of interest and influence the evolutionary process by selecting the favorite solution displayed in the MAP-Elites grid.

\subsection{Discussion}\label{sec:discussion}
%\todo[inline,color=green]{Antonios reviewed}

%constraints
Observing the cases of PCG-QD in Table~\ref{tab:cases}, most cases analyzed use constrained optimization to ensure playable content. This is likely due to games being an \emph{ergodic medium}~\cite{aarseth1997cybertext}, which cannot be used if the generated content is broken. 
Since assessing playability often requires extensive tests (e.g. via simulations), constrained optimization approaches are particularly well suited for this. In PCG-QD, the quality component satisfies the feasibility constraints, and is enhanced by the search for diverse content performed within the constrained search space. Future research should enhance PCG-QD with a quality component among only feasible individuals, as in~\cite{khalifa2018talakat}.
%Specifically, we can notice that four of them have constraints to ensure playable content, while only one has aesthetic constraints.
%quality can be defined based on the required constraints, such as playability, and then search for diverse solutions within the constrained search space.

%used why in the listed cases
It is also interesting to inspect the motivation for PCG-QD (see Section \ref{sec:why_qd}) used in the surveyed work of Table \ref{tab:cases}. 
\emph{Generative efficiency} and \emph{Human-machine Co-creation} are featured in~\cite{liapis2013sentient}: the former is beneficial for generating multiple suggestions simultaneously in quasi-real-time\footnote{Sentient Sketchbook~\cite{liapis2013sentient} also uses standard SBPCG  to generate only one high-quality suggestion per evolutionary run, requiring 6 threads for 6 suggestions while CNS produces 6 suggestions in 1 thread.}, while the latter allows the tool to assist the user with both diverse and good suggestions.
\emph{Fitness-free search} motivates~\cite{gravina2016constrained} to generate weapons with unexpected uses, such as the mine-layer. Regarding \emph{online expressivity} analysis, we may argue that it covers (in some cases implicitly) all the described cases of PCG-QD.
For instance, in \cite{fontaine2019mapping} the diversity characterization is used to inspect the expressive features of the cards evolved. More importantly, the underlying distribution of the cards' mana plays an active role during the evolutionary search by affecting the boundaries of the space partitioning.
Finally, \emph{explainability} to the designer was implicitly provided in EDD \cite{alvarez2019empowering}; however, to the best of our knowledge, no project has explicitly targeted explainability in the cases surveyed, which hints at a possible direction for future work.

\section{Open problems and Outlook}~\label{sec:outlook}

While several instances of PCG-QD have explored the space of possibilities in games and beyond (surveyed in Section \ref{sec:cases}), there is extensive work to be done in this direction. We highlight some of the challenges and some of the most promising avenues for exploring QD approaches in games.

A core challenge of PCG-QD is the definition of an effective behavior characterization.
The first issue relates to the curse of dimensionality, as a multi-dimensional BC can curtail the advantages described in Section~\ref{sec:why_qd}. Exploring diverse solutions in high-dimensional spaces can hinder the performance of the QD algorithm. Specifically, in high-dimensional spaces it can be hard to find \emph{interesting} diversity across all the possible solutions of the search space~\cite{nguyen2015innovation}.
A related challenge is the chosen BC partitioning, as the selected granularity may affect the efficiency of the QD approach. A fine-grained partition can make coverage of the space difficult, while a coarse-grained partition might hide interesting solutions.  
Several ways to address this have been put forward, such as dimensionality reduction in latent spaces~\cite{liapis2013transforming} or using centroidal Voronoi tessellation to maintain a constant partition of the space regardless of the number of dimensions~\cite{vassiliades2018using}. However, future work is required to investigate further how to best handle high-dimensional states.
Beyond quality-diversity, an interesting direction for future work would be to fuse different PCG approaches, such as PCGML and PCG-QD.
As we highlighted in Section~\ref{sec:cases}, some early examples of a fusion between quality-diversity and machine learning have been tested already~\cite{liapis2013transforming, nguyen2015innovation}. 
However, an interesting future direction would be to exploit the unique capabilities of PCGML (e.g., autonomous generation, data compression and analysis of the content~\cite{summerville2018procedural}) with the advantages of quality-diversity (see Section~\ref{sec:why_qd}).
For instance, machine-learned models of gameplay \cite{karavolos2018surrogate} could be used to improve the computational efficiency of simulation-based PCG-QD solutions. Numerous and long evaluations of content are usually required to test the quality of the generated content both in SBPCG and PCG-QD. However, we can imagine leveraging the abstraction capabilities of the machine-learned model to create surrogate of the simulations; this approach can reach comparable results in fewer evaluations~\cite{gaier2018data}.

Another direction for future work in games is applying QD algorithms to debug, diagnose and understand game playing agents. QD algorithms can generate a diverse set of levels that capture different important features of the level itself and an agent's gameplay~\cite{fontaine2019mapping,khalifa2019intentional}. By analyzing the outputs, we can understand which behavior has the most effect on the performance of the playing agent. The same idea can be used to improve the generality of reinforcement learning algorithms~\cite{justesen2018illuminating} by generating diverse sets of levels that can be validated to be playable but the agent can't beat them \cite{anderson2018deceptive}. %A similar approach is followed by the Paired Open-Ended Trailblazer (POET) where environments are co-evolved with agents trying to navigate them.

%\todo[inline]{Daniele: add POET here ?}

%As noted in Section~\ref{sec:beyond}, 
Beyond procedural content generation,
QD can also be used to evolve a group of playing agents that are either different from each other or compatible with each other. This is especially important for agents playing in collaborative games, as we might need a group of agents that are different from each other and at the same time compatible. 
For instance, MAP-Elites was used to generate a diverse set of high quality agents~\cite{canaan2019hanabi} that play the game \emph{Hanabi} (Antoine Bauza, 2010). 
Another use for high-quality diverse agents is for testing generated content with different play styles~\cite{liapis2015personacritics}. This can help debug the game to see if the current content is experienced in a similar manner as intended.

While the cases surveyed in Section~\ref{sec:cases} focused mostly on the generation of levels and visuals, it is important to explore how QD algorithms can work with different facets of game content such as rules~\cite{khalifa2017general}, music~\cite{hutchings2019adaptive}, etc. These facets come with their own challenges in defining quality or diversity. For example, in rule generation it is easy to define playable games (games that an automated agent can win in a number of steps) but a formula for good games is more difficult to devise. QD approaches can help explore the space of acceptable artifacts with different characteristics, allowing us to understand the effect these characteristics have on the generated artifacts.

\section{Conclusion}\label{sec:conclusion}

In this paper, we distinguished quality-diversity (QD) as a search strategy for search-based procedural content generation. Based on a range of recent applications of QD to games, QD algorithms can produce a large set of diverse (through controllable and often designer-friendly dimensions) and high-quality content (through constraints on playability and/or local competition). This makes PCG-QD particularly efficient in producing many diverse artifacts in one run, which is useful as explainable designer feedback, in a mixed-initiative tool or for expressivity analysis. Based on the current work in this vein, we identified under-explored areas in terms of algorithms and intended uses. Finally, we laid out a vision for the future of the field and the challenges that it will have to overcome.

%In this paper, we defined how the paradigm of quality-diversity can be applied to Procedural Content Generation. We motivated how generating a collection of diverse and high-quality content can be beneficial in this field, and surveyed existing previous work using this approach. Along with constructive, SBPCG and PCGML, we argue that PCG-QD is a useful tool to generate diverse and creative content.

\section*{Acknowledgment}
This project has received funding from the European Union's Horizon 2020 programme under grant agreement No 787476.
Ahmed Khalifa acknowledges the financial support from NSF grant (Award number 1717324 - ``RI: Small: General Intelligence through Algorithm Invention and Selection.'').

\bibliographystyle{IEEEtran}
\bibliography{quality_diversity_pcg}

% that's all folks
\end{document}